%% file: AT-NMF_double_column_rev.tex
\documentclass[journal]{IEEEtran}

\ifCLASSINFOpdf
\else
   \usepackage[dvips]{graphicx}
\fi
\usepackage{url}
\usepackage{amsfonts}
\usepackage{graphicx,epstopdf}
\usepackage{color}
\usepackage{graphicx,bbm}
\usepackage{multirow}
\usepackage{subcaption}
\usepackage{comment}
\usepackage[linesnumbered,boxed,ruled,commentsnumbered]{algorithm2e}
\usepackage{adjustbox}
\usepackage{graphicx, subcaption}
\usepackage{algorithmic}

\usepackage{grffile}

\let\oldnl\nl
\newcommand{\nonl}{\renewcommand{\nl}{\let\nl\oldnl}}

\usepackage{amsmath}

\DeclareMathOperator*{\argmin}{arg\,min}

\input{preamble}
\begin{document}

\title{Adversarially-Trained Nonnegative Matrix Factorization}

\author{Ting Cai,  Vincent Y. F.  Tan, \IEEEmembership{Senior Member, IEEE},  and C\'edric F\'evotte, \IEEEmembership{Senior Member, IEEE}\thanks{T.~Cai and V.~Y.~F.~Tan are with the   National University of Singapore (emails: elect@nus.edu.sg, vtan@nus.edu.sg).  C. F\'evotte is with IRIT, Universit\'e de Toulouse, CNRS, Toulouse, France (email: cedric.fevotte@irit.fr). This work is funded by a Singapore NRF Fellowship (R-263-000-D02-281) and by the European Research Council (ERC FACTORY-CoG-6681839). }}
%

\markboth{Journal of \LaTeX\ Class Files, Vol. 14, No. 8, August 2015}
{Shell \MakeLowercase{\textit{et al.}}: Bare Demo of IEEEtran.cls for IEEE Journals}
\maketitle
 
\begin{abstract}
We consider an adversarially-trained version of the nonnegative matrix factorization, a popular latent dimensionality reduction technique. In our formulation, an attacker adds an arbitrary matrix of bounded norm to the given data matrix. We design efficient algorithms inspired by adversarial training to optimize for  dictionary and coefficient matrices with enhanced generalization abilities. Extensive simulations on synthetic and benchmark datasets demonstrate the superior predictive performance on matrix completion tasks of our proposed method compared to state-of-the-art competitors, including other variants of adversarial nonnegative matrix factorization.
\end{abstract}


\IEEEpeerreviewmaketitle

\vspace{-.025in}
\section{Introduction}
\vspace{-.025in}
\IEEEPARstart{N}{onnegative} matrix factorization (NMF) is a dictionary learning technique for learning non-subtractive and parts-based representations of given nonnegative data \cite{LS99}. The mathematical formulation consists in decomposing a given nonnegative data matrix $\bV\in\bbR_+^{F\times N}$ into two nonnegative matrices---the dictionary matrix $\bW\in\bbR_+^{F\times K}$ and the coefficient matrix $\bH\in\bbR_+^{K\times N}$ such that $\bV\approx\bW\bH$. To find an approximate decomposition, one solves the non-convex problem  (known throughout the paper as {\em standard NMF})
\begin{align}
	\min_{\bW, \bH\ge \bzero} \, \| \bV - \bW\bH  \|^2_{\rmF}, \label{eqn:standardNMF}
\end{align}
where $\bA \ge \bzero$ means all entries of the matrix $\bA$ are nonnegative. Popularized by Lee and Seung~\cite{LS99}, NMF has found a variety of applications in matrix completion~\cite{fevotte2011algorithms}, document clustering~\cite{xu2003document} and audio signal processing~\cite{fevotte2009nonnegative}. For a thorough review on NMF, the reader is referred to the book by Gillis~\cite{gillisBook}. 

As demonstrated in the landmark work by Goodfellow {\em et al.} \cite{goodfellow2015explaining}, machine learning methods are vulnerable to adversarial attacks on the observed data. To mitigate this and to improve generalization, adversarial training (AT) has recently come to the fore. AT has been shown to improve the robustness of deep learning models in~\cite{madry2017towards}  and~\cite{tramer2018ensemble}. More recently, Sinha {\em et al.}~\cite{sinhaND18} and Farnia {\em et al.}~\cite{farniaAT} showed theoretically that AT improves generalization in generic machine learning tasks. 

In this paper, we seek to improve the predictive performance of NMF on matrix completion tasks by considering an adversarially-trained version of the standard NMF objective in~\eqref{eqn:standardNMF}. While there is a previous work by Luo {\em et~al.}~\cite{pmlr-v119-luo20c}, called {\em adversarial NMF} or {\em ANMF}, which attempts to address a similar problem, here we highlight some salient differences between \cite{pmlr-v119-luo20c} and our formulation.  Luo {\em et al.}~\cite{pmlr-v119-luo20c} applied  adversarial training {\em only} to the dictionary matrix $\bW$. Their formulation necessitates the introduction of a so-called {\em instance-specific target} $\bZ$ (a hyperparameter) for the adversary; this target $\bZ$ is not easy to choose in a principled way in practice.  The adversary learns a dictionary matrix $\tilbW$ such that $\tilbW\bH$ is close to $\bZ$. The learner then optimizes a linear combination of two objectives---one objective encouraging $\tilbW \bH$ to be close to the data matrix $\bV$ and another encouraging $\bW\bH$ to be close to $\bV$. The authors
used an ADMM algorithm for this bilevel optimization and demonstrated that ANMF is able to increase the robustness of NMF model on  clustering tasks.  We consider the case in which the attacker is allowed  to attack  $\bV$ without being instance-specific, i.e., we consider the problem of minimizing $(\bW,\bH)\mapsto\max_{\bR}\|\bV + \bR -\bW\bH\|_\rmF^2$ where $\bR$ belongs to a bounded set $\calR$. In contrast to ANMF, our formulation has only {\em one scalar} hyperparameter $\lambda$. We derive efficient updates for $\bR$ and $(\bW,\bH)$, which is known as {\em adversarially-trained NMF} or {\em AT-NMF}. We  compare the prediction performance of AT-NMF to ANMF and standard NMF and show the superior performance on matrix completion tasks for three benchmark datasets  commonly used in NMF.

%

%
%
%
%
\vspace{-.025in}
\section{Adversarially-Trained NMF} \label{sec:atnmf}\vspace{-.025in}
We  now  introduce the formulation for AT-NMF. There exists an adversarial attacker who adds an {\em arbitrary} matrix $\bR \in \mathbb{R}^{F \times N}$  to the data matrix~$\bV$. The {\em adversarial matrix} $\bR$  maximizes the Frobenius norm 
between  $\bV$ and our desired factorization $\bW\bH$; see the similar manner in which   adversarial noise is added on to the training data in~\cite{sinhaND18, farniaAT}.  Since the adversary typically has limited power, we assume that $\bR$ belongs to a bounded set. Thus, AT-NMF can be formulated~as 
\begin{align}
  \min_{\bW,\bH \ge \bzero}\, \max_{\bR \in \mathcal{R}}\, \| \bV+\bR - \mathbf{WH} \|^2_{\rmF} \label{eqn:atnmf}
  \end{align}
 where the constraint set
$
 \mathcal{R} \! := \!\big\{ \bR: \|\bR \|^2_{\rmF} \!\le\! \epsilon, \bV \!+\! \bR\! \ge\! \bzero \big\}.$
 Here, $\epsilon>0$ is a constant that indicates the adversary's power; the larger $\epsilon$ is, the larger the adversary's power and {\em vice versa}.  We assume that $\bR$ belongs to the set of Frobenius norm-bounded matrices and $\bV+\bR$ is mandated to be nonnegative so that for any fixed $\bR$, the optimization in~\eqref{eqn:atnmf} corresponds to a standard NMF problem with effective data matrix $\bV+\bR$.  We note that a similar  min-max formulation has appeared in the recent literature~\cite{gillisRobust} but the objective therein is different as the authors sought to only be robust to the $\beta$-divergence measure used and not perturbations to the  data matrix $\bV$.

The problem as stated in \eqref{eqn:atnmf}  is difficult to solve in general. Hence, we resort to a relaxation inspired by Lagrangian duality~\cite{Boyd04}. We dualize the norm constraint $\|\bR \|^2_{\rmF} \le \epsilon$  with a Lagrange multiplier $\lambda>0$ so that the optimization problem becomes 
\begin{equation}
  \min_{\bW,\bH \ge \bzero}\, \max_{\bR : \bV+\bR\ge\bzero}\, \| \bV+\bR - \bW\bH  \|^2_{\rmF}  - \lambda \|\bR\|_\rmF^2. \label{eqn:minmax_lambda}
\end{equation}
Note that the regularization term has a {\em negative} coefficient because the inner optimization in \eqref{eqn:minmax_lambda} is a {\em maximization}.  The interpretation of $\lambda$ is   dual to that of $\epsilon$; it indicates the adversary's power. Indeed as $\lambda\to0^+$, we see that the norm of $\bR$ will be unbounded, indicating an extremely powerful adversary. Conversely, as $\lambda\to\infty$, the effect of the regularization term vanishes and the norm of $\bR$ tends to zero, indicating that there is effectively no adversary. 

The optimization problem in \eqref{eqn:minmax_lambda} can naturally be broken up into two parts, an {\em inner} maximization  in which we optimize over $\bR$ and an {\em outer} minimization  in which we optimize over $(\bW,\bH)$. The inner maximization  can be  equivalently rewritten as a minimization problem as
\begin{equation}
\bR^*=\argmin_{\bR: \bV+\bR\ge\bzero} \, -\| \bV+\bR - \mathbf{WH} \|^2_{\rmF}  + \lambda \|\bR \|_\rmF^2.\label{eqn:minR}
\end{equation}
We show in the next section that the solution to this minimization  problem is well-defined for $\lambda>1$; indeed, it can be found in closed-form. After $\bR^*$ has been found, one can then aim to minimize
\begin{equation}
\| \bV+\bR^* - \bW\bH  \|^2_{\rmF}  \label{eqn:minWH}
\end{equation}
over $\bW,\bH\ge \bzero$. By treating $\bV+\bR^*$ as a new data matrix, we can use Majorization-Minimization (MM) \cite{Hunter} to find approximate solutions for $\bW$ and $\bH$. We note that since we are using the Frobenius norm, there are alternatives to MM.

%
%

\subsection{Update of $\bR$ in \eqref{eqn:minR}}

We now provide details on how to solve for $\bR^*$  in \eqref{eqn:minR}. Here the matrices $\bW$ and $\bH$ are fixed and we denote their product as $\hatbV=\bW\bH$.   Thus, we can denote the objective as 
\begin{equation}
g(\bR):=- \|\bV+\bR -\hatbV\|_\rmF^2 +\lambda \|\bR\|_\rmF^2,  \label{eqn:gFunc}
\end{equation}
where $\bR$ is constrained such that $\bV+\bR\ge\bzero$. 
We note that this objective, by virtue of being the sum of two squared Frobenius norms,  is {\em separable}, i.e., it decomposes into the sum of $FN$ independent terms, i.e., 
\begin{equation}
g(\bR) =  \sum_{f=1}^F\sum_{n=1}^N \left[-(v_{fn}+r_{fn} - \hatv_{fn})^2+\lambda r_{fn}^2\right]
\end{equation}
Thus, to minimize $g(\bR)$ over  $\bR$, it suffices to minimize each of the constituent terms in the sum over $r_{fn}$. In the following, we abbreviate $v_{fn}$, $r_{fn}$ and $\hatv_{fn}$ to $v$, $r$ and $\hatv$ respectively.

The problem thus reduces to the scalar optimization 
\begin{equation}
\min_{r: v+r\ge 0 } \, -(v+r  - \hatv)^2+\lambda r^2
\end{equation}
which is equivalent to 
\begin{equation}
\min_{r:v+r \ge 0} \,(\lambda - 1)r^2 - 2r(v-\hatv) .  \label{eqn:single}
\end{equation}
If $0\le \lambda < 1$, the objective function is strictly concave in~$r$. Thus the optimal value of $r$ is $ \infty$, which indicates that the adversary has unbounded power, an unrealistic scenario. When $\lambda=1$, we again have undesirable solutions. Indeed, the objective function in \eqref{eqn:single} is linear. If $\hatv-v > 0$, then the solution is $r^* =-v$; if $\hatv= v$, regardless of the value of $r$, the   function is $0$; finally, if $\hatv-v<0$, the solution is $r^*=\infty$.  To avoid handling these cases separately, we avoid using $\lambda=1$. 

Thus the only meaningful case, and the case we consider in our experiments, is   $\lambda>1$. In this case, the objective function in~\eqref{eqn:single} is strictly convex and it is easy to show by direct differentiation that the optimal constrained solution to~\eqref{eqn:gFunc} under the condition that $\bV+\bR\ge \bzero$ is  
\begin{equation}
\bR=\max\left\{ \frac{\bV-\hatbV}{\lambda-1} ,-\bV\right\} ,\label{eqn:updateR}
\end{equation}
where the maximization operator is applied elementwise.  

\subsection{Update of $\bW$ and $\bH$ in \eqref{eqn:minWH}}
Upon the optimization of $\bR$ via the formula in~\eqref{eqn:updateR}, the optimization over $(\bW,\bH)$ is standard if we regard $\bU:=\bV+\bR^*$ as the {\em effective data matrix}. The classical MM steps~\cite{fevotte2011algorithms,Hunter} lead to multiplicative updates for $\bW$ and $\bH$ as follows
\begin{equation}
\bH \leftarrow \bH \cdot  \frac{\bW^\top\bU}{\bW^\top\bW\bH } \quad\mbox{and}\quad \bW \leftarrow \bW \cdot \frac{\bU\bH^\top}{\bW\bH \bH^\top}, \label{eqn:updateWH}
\end{equation}
where $\cdot$ and $\cdot/\cdot$ respectively denote elementwise multiplication and elementwise division respectively.  

Just to recapitulate, we first update $\bR$ via \eqref{eqn:updateR} and then update $(\bW,\bH)$ via \eqref{eqn:updateWH}. These steps are shown in Algorithm~\ref{alg:atnmf}.

\subsection{Initialization of $(\bW,\bH)$} \label{sec:init}



We now describe how we initialize the dictionary and coefficient matrices $\bW_{\mathrm{init}}$ and $\bH_{\mathrm{init}}$. First, we sample each entry of each matrix independently from a Half-Normal distribution~\cite{Leone}  with precision parameter $\gamma=1$. Then, starting with these matrices, we run $5$ standard MM steps according to~\eqref{eqn:updateWH} on the given data matrix $\bV$  to obtain $\bW_{\mathrm{init}}$ and $\bH_{\mathrm{init}}$. The reason why we use  this initialization method is to get a pair $( \bW_{\mathrm{init}},   \bH_{\mathrm{init}})$ such that $\hat{\bV}_{\mathrm{init}}   =\bW_{\mathrm{init}}   \bH_{\mathrm{init}} \approx\bV$. Otherwise, as~\eqref{eqn:updateR} shows,  $\bR^*$ may be set to $-\bV$ so that $\bU=\bV+\bR^*=\bzero$, which renders the subsequent steps vacuous. This initialization is computationally cheap and effective. 

\subsection{Stopping Criteria of AT-NMF}
Since $\bR$ can be found in closed-form if $\lambda>1$, no termination criterion for updating $\bR$ is required. However, the optimization of $(\bW,\bH)$ is found via iterative optimization according to~\eqref{eqn:updateWH}. So, in this subsection, we discuss the criterion we employ to terminate this inner optimization before reverting to computing a new $\bR$. We consider the {\em relative error} of successive iterates and terminate once this error falls below a certain threshold $\varepsilon_{\mathrm{in}}$, i.e., if $(\bW^{(o, i)} , \bH^{(o, i)})$ denotes the iterate of $(\bW,\bH)$ at the $o^{\text{th}}$ outer iteration and $i^{\text{th}}$ inner iteration and $\hatbV^{(o,i)}:=\bW^{(o,i)}\bH^{(o,i)}$, we terminate the inner optimization once the inner iteration index   $i$ satisfies
\begin{equation}
\left\| \frac{\hatbV^{(o,i+1)}-\hatbV^{(o,i)}}{\hatbV^{(o,i)}} \right\|_\rmF<\varepsilon_{\mathrm{in}}. \label{eqn:inner_cond}
\end{equation}
for some $\varepsilon_{\mathrm{in}}>0$.
The entire optimization is terminated  once the outer iteration index $o$ satisfies 
\begin{equation}
\left\| \frac{\hatbV^{(o+1,i)}-\hatbV^{(o,i)}}{\hatbV^{(o,i)}} \right\|_\rmF<\varepsilon_{\mathrm{out}}  \label{eqn:outer_cond}
\end{equation}
for some $\varepsilon_{\mathrm{out}}>0$ where  $i$ is the index  of the inner iteration for which we terminated previously. In practice, we also terminate the inner and outer loops once preset number of iterations \texttt{maxInner} and \texttt{maxOuter} respectively are reached. 
\begin{algorithm}[ht] 
	\caption{AT-NMF$(\bV,\lambda,  K)$ } \label{alg:atnmf}
		\begin{algorithmic}[1]
			\STATE {\bfseries Input:} Data matrix $\bV\in\bbR_+^{F\times N}$; Regularization param.~$\lambda$; Stopping params.~$\varepsilon_{\mathrm{in}}$ and~$\varepsilon_{\mathrm{out}}$; Common dim.~$K$; Integers \texttt{maxInner} and \texttt{maxOuter};
			\STATE {\bfseries Initialize:} $(\bW,\bH)$ to $(\bW_{\mathrm{init}},\bH_{\mathrm{init}})$  per Sec.~\ref{sec:init}.
			\STATE \texttt{outerNotConv} $=$ {\sc True};  
			\WHILE{\texttt{outerNotConv}}
                \STATE Update $\bR$ using \eqref{eqn:updateR};
                \STATE \texttt{innerNotConv}  $=$ {\sc True}
			    \WHILE{\texttt{innerNotConv}}
			        \STATE Update $(\bW,\bH)$ using \eqref{eqn:updateWH};
			        \IF{Cond.\ \eqref{eqn:inner_cond} holds or \texttt{maxInner} reached}
		        	\STATE \texttt{innerNotConv} $=$ {\sc False}
			        \ENDIF	
			    \ENDWHILE			
			    \IF{Cond.\ \eqref{eqn:outer_cond} holds or \texttt{maxOuter} reached}
			        	\STATE \texttt{outerNotConv} $=$ {\sc False}
			    \ENDIF  
  	       \ENDWHILE	
		\end{algorithmic}
	\end{algorithm}
	\vspace{-.1in}

\section{Numerical Experiments} \label{sec:numerical}

We perform   numerical experiments on synthetic and real datasets to demonstrate the prediction and generalization efficacy of AT-NMF vis-\`a-vis standard NMF~\cite{LS99} and ANMF~\cite{pmlr-v119-luo20c}. The hyperparameters that we use   are $\texttt{maxInner} = 1000$, $\texttt{maxOuter} = 100$, $\varepsilon_{\mathrm{in}}=\varepsilon_{\mathrm{out}}=0.01$. Our experiments all involve predicting entries of $\bV$ that are held out. We denote the fraction of held-out entries as $\alpha\in (0,1)$ which takes on values in  $\{0.1, 0.2,\ldots, 0.8,0.9\}$. That is, for $\alpha=0.9$, we randomly remove 90\% of the entries of $\bV$, train the model  (on standard NMF, ANMF and AT-NMF) on the remaining 10\% of available entries and assess the prediction performance thereafter.   
Like standard NMF, Algorithm \ref{alg:atnmf} can straightforwardly be adapted to handle missing values by applying a binary mask to $\bV$, see, e.g., \cite{fevotte2011algorithms}. Note that $r_{fn}$ is only computed for observed  values.
 To be fair to all competing algorithms, we use the same stopping criterion as AT-NMF for NMF and the default parameters for  ANMF \cite{pmlr-v119-luo20c}.  The code to reproduce   our experiments can be found at \url{https://github.com/caiting123321/AT_NMF}.

\begin{table}[t]
\begin{center}
\caption{RMSE of the synthetic dataset (Numbers in the brackets indicate the $\lambda$ used in AT-NMF)}
\label{tab:synthetic}
\begin{adjustbox}{width=.5\textwidth}
\begin{tabular}{ |c|c|c|c|c|c| } 
\hline
$\alpha$ & \textbf{NMF} & \textbf{ANMF} & \textbf{AT-NMF} ($2$) & \textbf{AT-NMF} ($3$) & \textbf{AT-NMF} ($5$)  \\
\hline
$0.3$ & $5.37 \pm 0.02$ & $6.78 \pm 0.17$ & $5.41 \pm 0.12$  & $\mathbf{5.11} \pm \mathbf{0.03}$ & $5.20 \pm 0.02$   \\ 
\hline
$0.4$ & $5.62 \pm 0.03$ & $6.92 \pm 0.17$ & $5.54 \pm 0.08$  & $\mathbf{5.32} \pm \mathbf{0.09}$ & $5.42 \pm 0.04$ \\ 
\hline
$0.5$ & $6.41 \pm 0.01$ & $7.44 \pm 0.09$ & $6.27 \pm 0.11$  & $\mathbf{6.05} \pm \mathbf{0.03}$ & $6.18 \pm 0.02$  \\ 
\hline
$0.6$ & $6.74 \pm 0.02$ & $7.61 \pm 0.09$ & $6.47 \pm 0.07$  & $\mathbf{6.39} \pm \mathbf{0.03}$ & $6.53 \pm 0.02$  \\ 
\hline
$0.7$ & $7.30 \pm 0.01$ & $7.99 \pm 0.06$ & $7.02 \pm 0.04$  & $\mathbf{6.94} \pm \mathbf{0.01}$ & $7.10 \pm 0.02$   \\ 
\hline
$0.8$ & $7.87 \pm 0.01$ & $8.30 \pm 0.06$ & $7.69 \pm 0.04$  & $\mathbf{7.61} \pm \mathbf{0.03}$ & $7.71 \pm 0.00$   \\ 
\hline
$0.9$ & $8.45 \pm 0.01$ & $8.58 \pm 0.06$ & $8.44 \pm 0.02$  & $\mathbf{8.34} \pm \mathbf{0.02}$ & $8.35 \pm 0.02$   \\ 
\hline
\end{tabular}
\end{adjustbox}
\end{center}
\end{table}

\begin{figure}[t]
\centering
\begin{tabular}{cc} 
	\hspace{-.1in}\includegraphics[width=0.46\columnwidth]{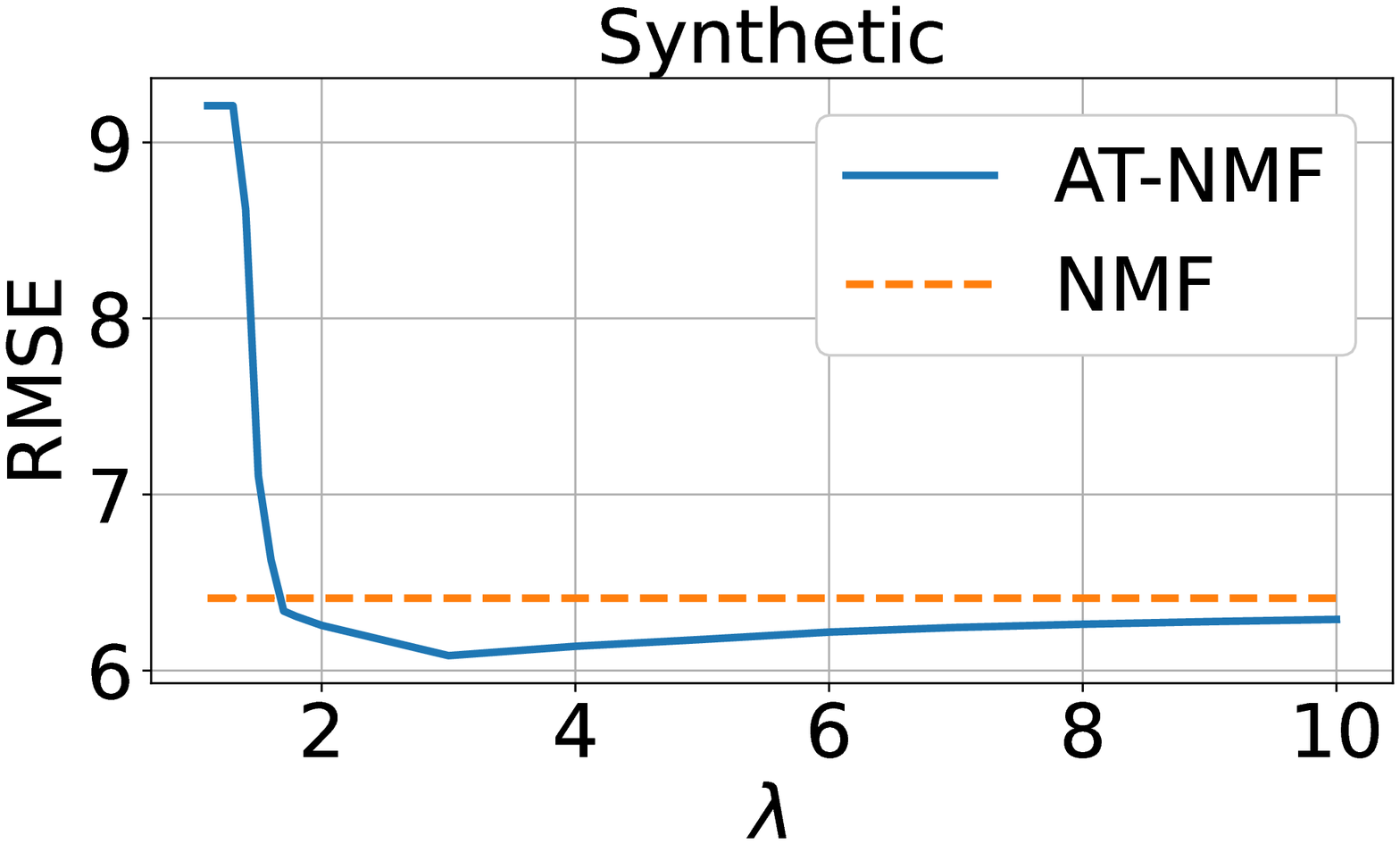} & \hspace{-.1in} \includegraphics[width=0.5\columnwidth]{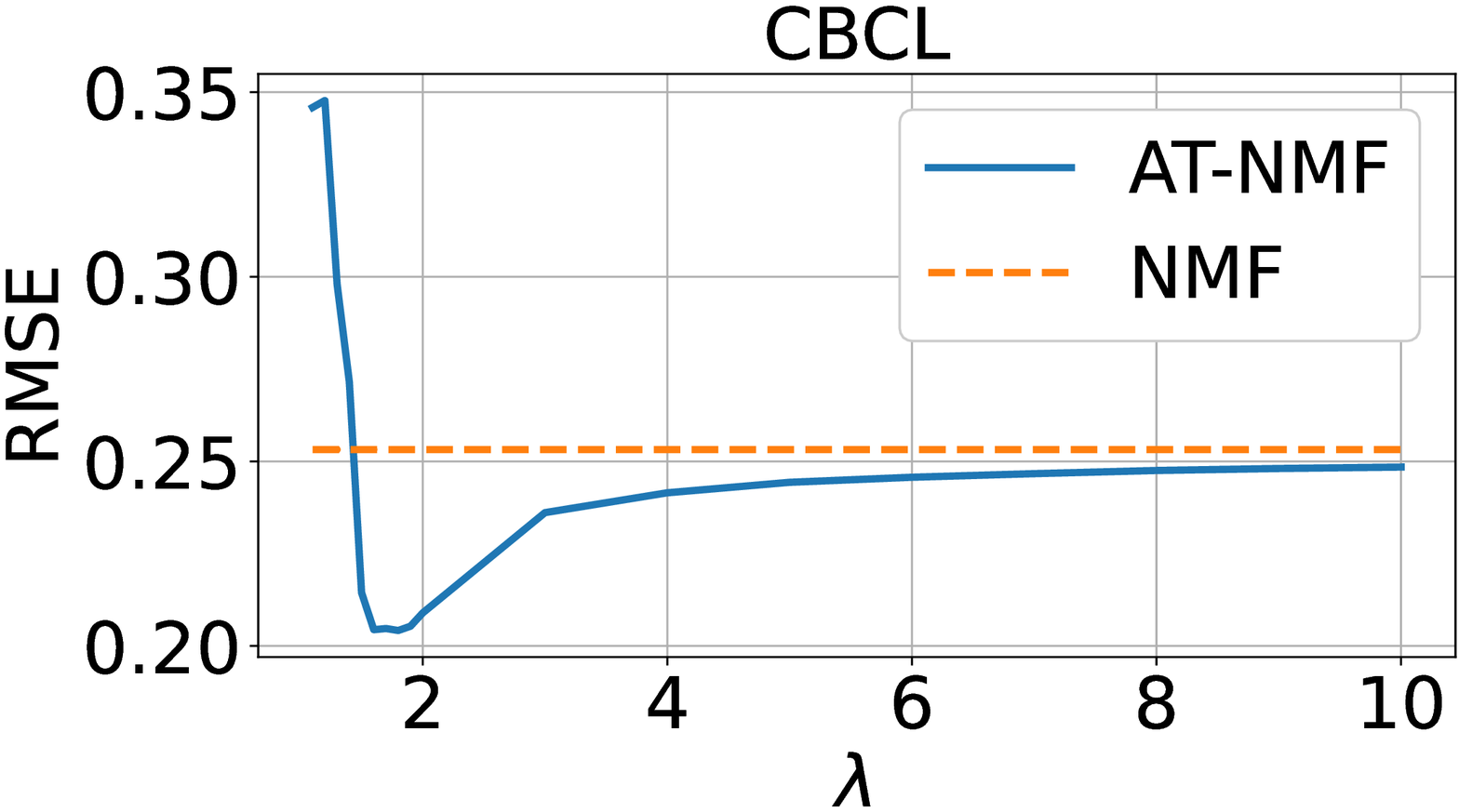}  \\
	\hspace{-.1in}\includegraphics[width=0.5\columnwidth]{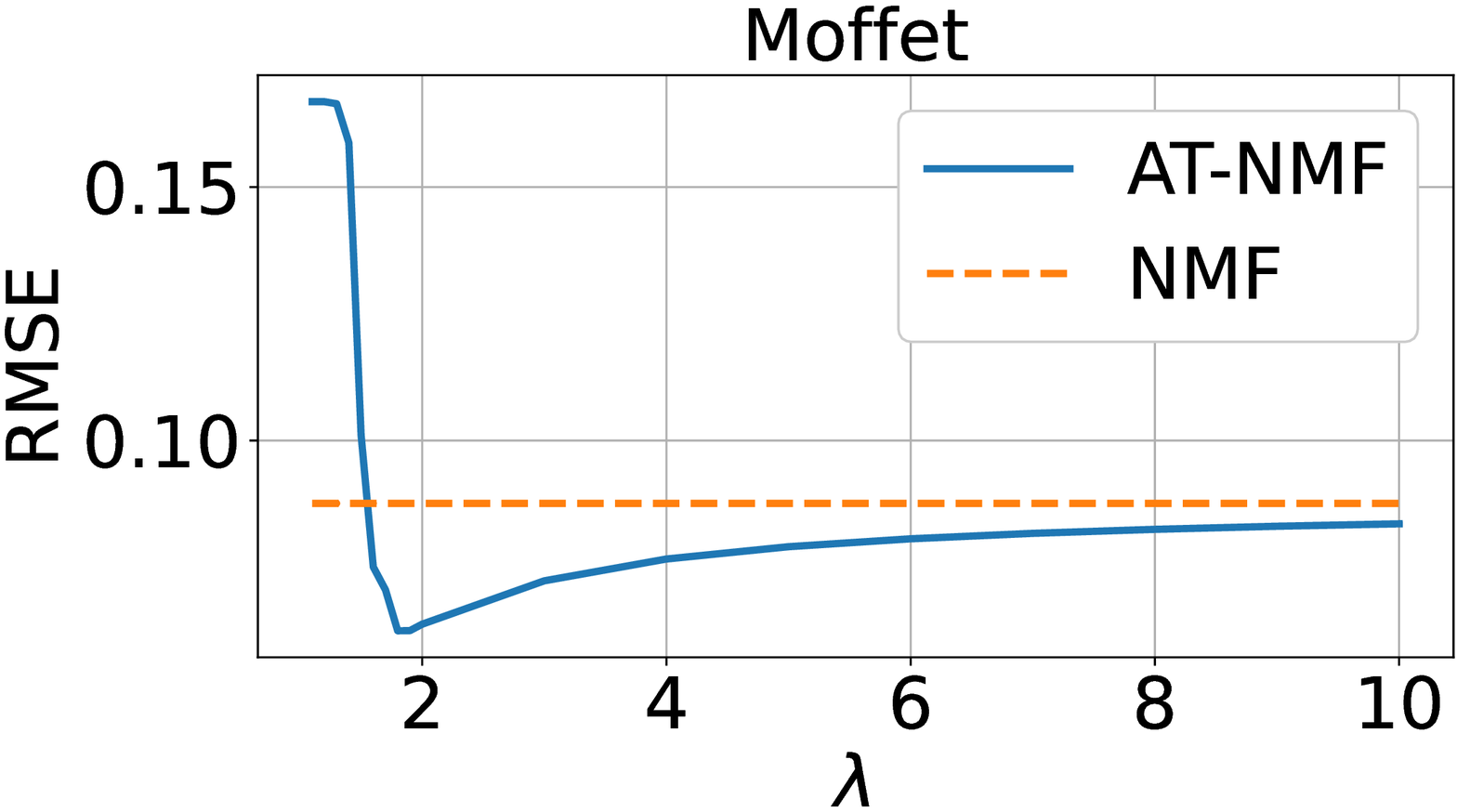} & \hspace{-.1in}
	\includegraphics[width=0.5\columnwidth]{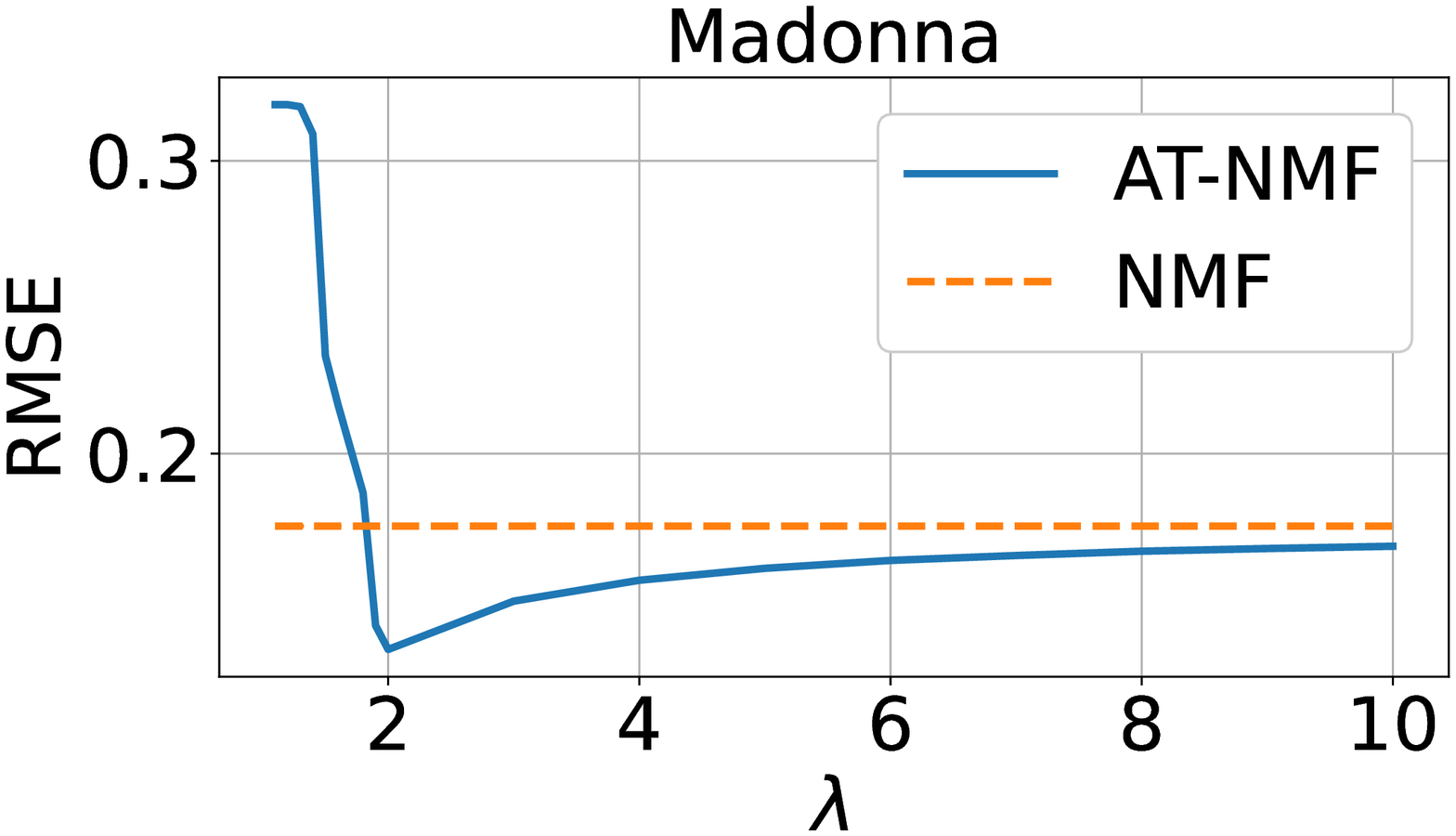}
\end{tabular}
\caption{Plots of $\mathrm{RMSE}$ against  $\lambda$ on the   datasets when $\alpha=0.5$}
\label{fig:rmse_lambda}
\end{figure}

Let $\Gamma\subset \{1,\ldots, F\}\times \{1,\ldots, N\}$ be the  set of held-out entries of $\bV$ so $|\Gamma|/(FN)=\alpha$. We denote the prediction of $\{v_{fn}: (f,n)\in\Gamma\}$ as $\{\hatv_{fn}: (f,n)\in\Gamma\}$ where $\hatv_{fn}=[\bW\bH]_{fn}$. Our performance metric is the {\em root mean-squared error} (RMSE), defined as 
\begin{equation}
\mathrm{RMSE}:=\sqrt{ \frac{1}{|\Gamma|}\sum_{(f,n)\in\Gamma} \big(v_{fn} -\hatv_{fn}\big)^2    }.
\end{equation}
The smaller the $\mathrm{RMSE}$, the better the prediction ability. All our results are averaged  over $10$ independent initializations (described in Sec.~\ref{sec:init}). In Tables~\ref{tab:synthetic}--\ref{tab:Madonna}, we report the standard deviations in addition to the means across the $10$ runs. 

%
%

\vspace{-.04in}
\subsection{Simulations on a Synthetic Dataset} \label{sec:synthetic}
\vspace{-.04in}
We describe experiments on synthetic data generated according to the model in \cite{TanFevotte2013}. We sampled $K = 5$ precision parameters $\gamma_k$ according to the inverse-Gamma distribution with parameters $(a=50, b=70)$. Conditioned on these precisions, we sampled the elements of the $k^{\text{th}}$ column of $\bW \in \bbR_+^{100 \times 5}$ and the $k^{\text{th}}$ row of  $\bH \in  \bbR_+^{5 \times 50}$ independently from the Half-Normal distribution with precision parameters $\gamma_k$ and $\bV =\bW \bH$. Our results are displayed in Table~\ref{tab:synthetic}.


First, we observe that all algorithms perform better when $\alpha$ is small. This is intuitive since there are more data to train on. More interestingly,  AT-NMF outperforms standard NMF and ANMF consistently over different $\alpha$  for the values of $\lambda$ chosen. For this dataset, ANMF performs worse than NMF. We believe that this because its hyperparameters, including the instance-specific target, have not been tuned (we are using the default ones). For this experiment, AT-NMF with  $\lambda=3$ outperforms AT-NMF with $\lambda=2$ and $\lambda=5$. Our final observation for the synthetic dataset pertains to scenario in which $\lambda$ becomes large. In this case, the adversary's power is diminished; hence, AT-NMF reverts to standard NMF. Indeed, the performance of AT-NMF for large $\lambda$ is close to  that of standard NMF. See the top left plot of Fig.~\ref{fig:rmse_lambda} where the $\mathrm{RMSE}$ of AT-NMF converges to that of NMF as $\lambda\to\infty$. This behavior was observed for the other real datasets as can be seen from the other subplots in Fig.~\ref{fig:rmse_lambda}. From our experience, the best choice of $\lambda$ depends on the average magnitude of the entries of $\bV$ and may require tuning to obtain the best performance.

%

\begin{table}[t]
\caption{RMSE of the CBCL   dataset}
\label{tab:CBCL}
\centering
\begin{adjustbox}{width=0.5\textwidth}
\begin{tabular}{ |c|c|c|c|c| } 
\hline
$\alpha$ & \textbf{NMF} & \textbf{ANMF} & \textbf{AT-NMF} ($2$)  & \textbf{AT-NMF} ($5$)  \\
\hline
$0.3$ & $0.204 \pm 0.001$ & $0.189 \pm 0.000$ & $\mathbf{0.176} \pm \mathbf{0.001}$ & $0.195 \pm 0.001$  \\ 
\hline
$0.4$ & $0.232\pm 0.001$ & $0.217 \pm 0.000$ & $\mathbf{0.192} \pm \mathbf{0.001}$ & $0.224 \pm 0.001$  \\ 
\hline
$0.5$ & $0.253 \pm 0.001$ & $0.241 \pm 0.000$ & $\mathbf{0.209} \pm \mathbf{0.001}$ & $0.244 \pm 0.001$  \\ 
\hline
$0.6$ & $0.272 \pm 0.000$ & $0.262 \pm 0.000$ & $\mathbf{0.227} \pm \mathbf{0.000}$ & $0.262 \pm 0.000$ \\ 
\hline
$0.7$ & $0.291 \pm 0.000$ & $0.285 \pm 0.000$ & $\mathbf{0.248} \pm \mathbf{0.001}$ & $0.283 \pm 0.000$  \\ 
\hline
$0.8$ & $0.309 \pm 0.000$ & $0.306 \pm 0.000$ & $\mathbf{0.277} \pm \mathbf{0.001}$ & $0.303 \pm 0.000$  \\ 
\hline
$0.9$ & $0.328 \pm 0.000$ & $0.326 \pm 0.000$ & $\mathbf{0.314} \pm \mathbf{0.000}$ & $0.324 \pm 0.000$  \\ 
\hline
\end{tabular}
\end{adjustbox}
\end{table}

\begin{figure}[hbt!]
\centering
\begin{tabular}{cc} 
\includegraphics[width=0.475\columnwidth]{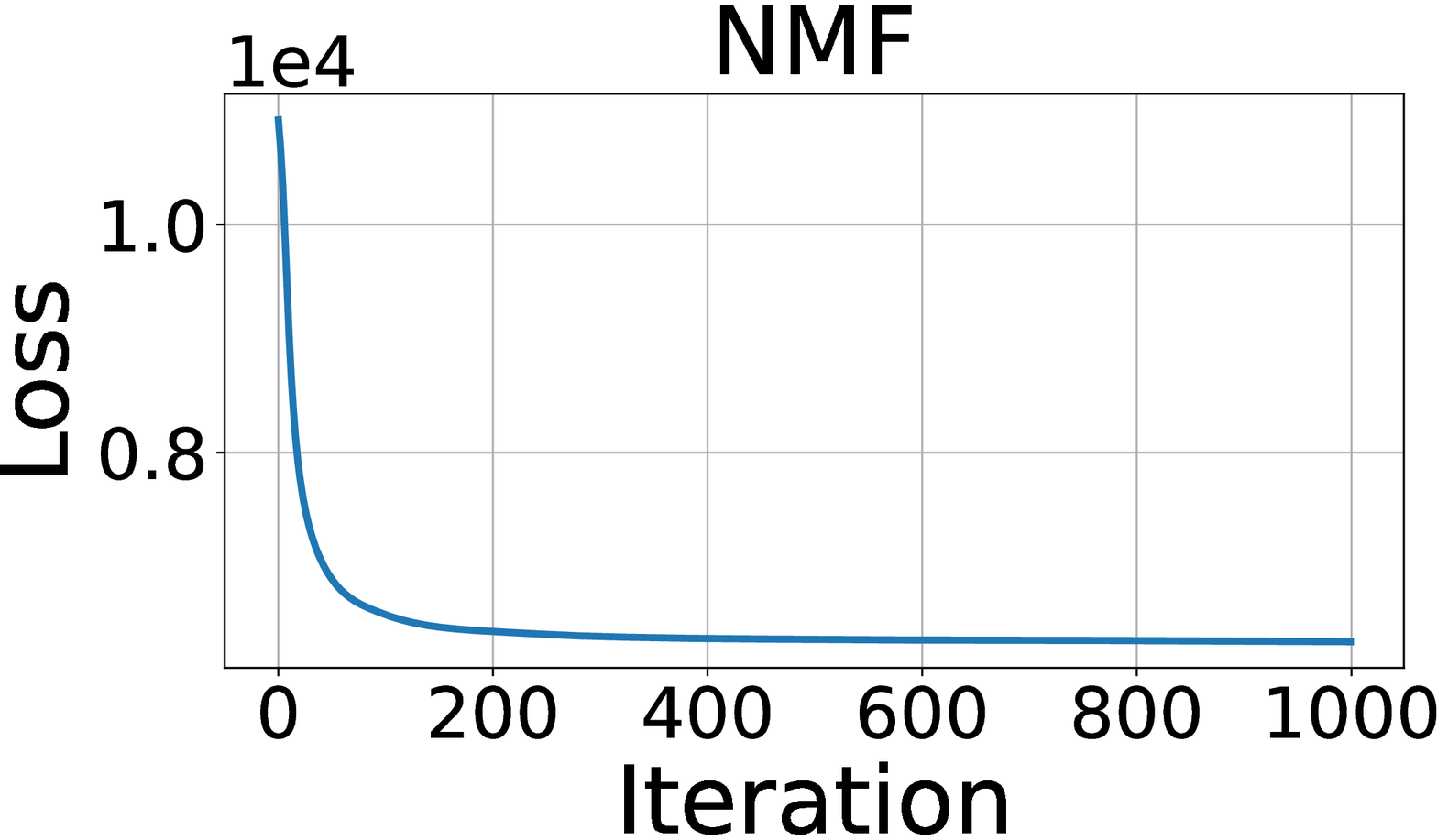} & \includegraphics[width=0.475\columnwidth]{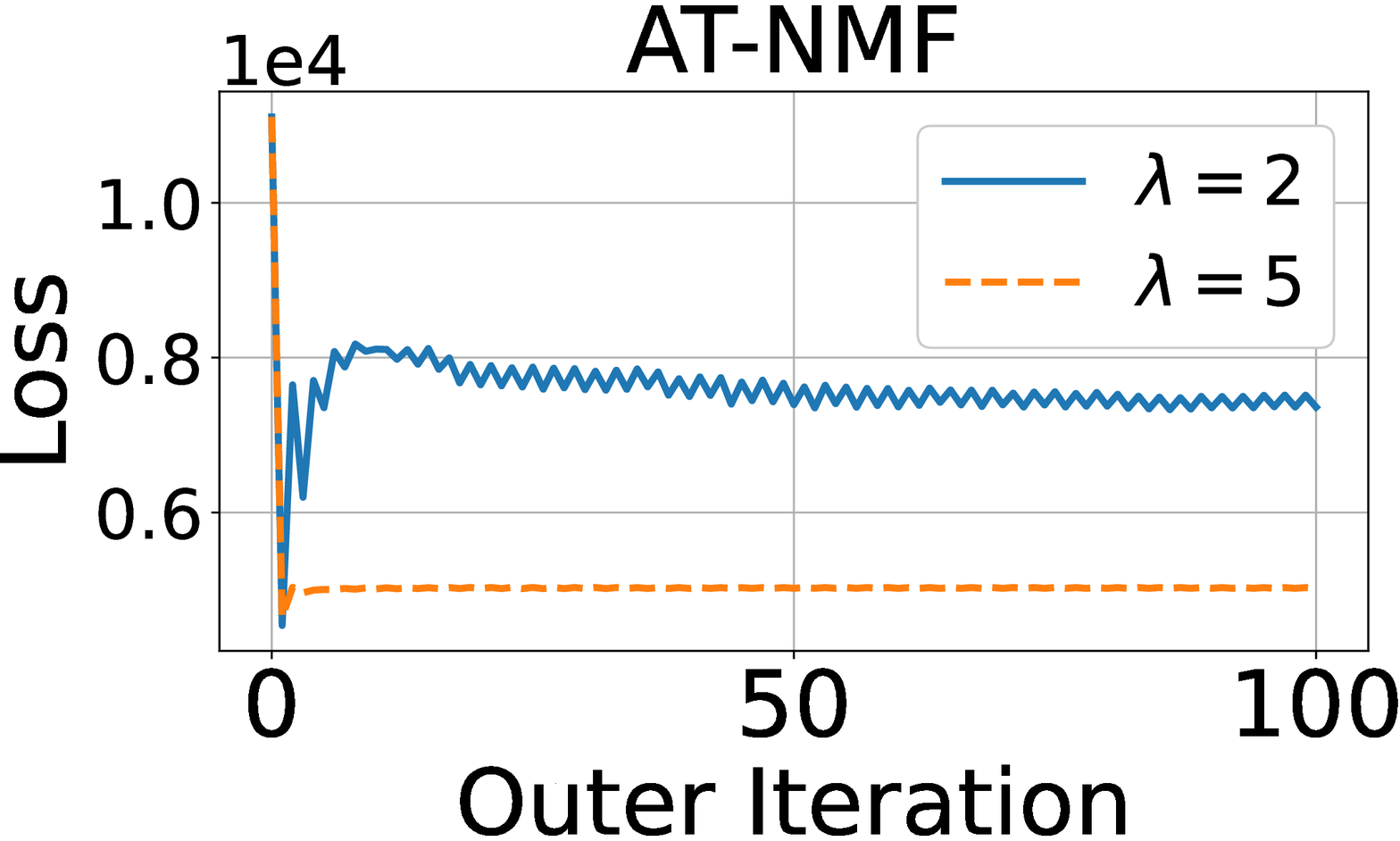}
\end{tabular}
\caption{Training losses of the CBCL dataset when $\alpha = 0.5$}
\label{fig:CBCL Loss}
\end{figure}

\begin{figure}[hbt!]
\centering
\begin{subfigure}[t]{.15\columnwidth}
	\centering
	\includegraphics[width = 1\columnwidth]{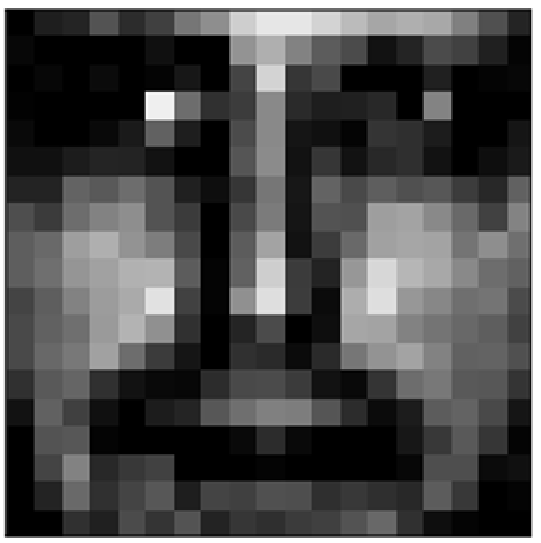}
	\subcaption{}
\end{subfigure}
\begin{subfigure}[t]{.15\columnwidth}
	\centering
	\includegraphics[width = 1\columnwidth]{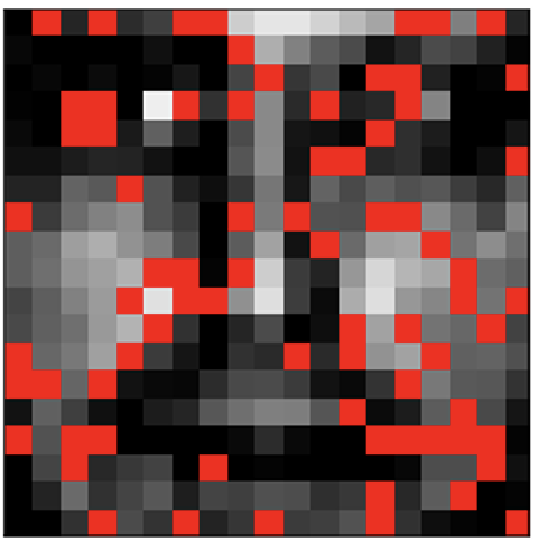}
	\subcaption{}
\end{subfigure}
\hfill
\begin{subfigure}[t]{.15\columnwidth}
	\centering
	\includegraphics[width = 1\columnwidth]{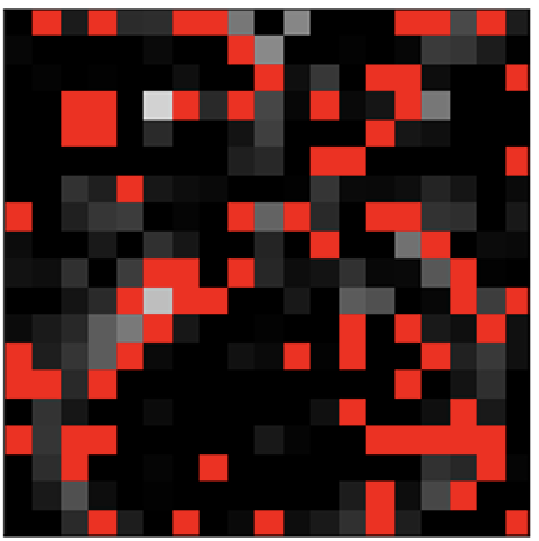}
	\subcaption{}
\end{subfigure}
\hfill
\begin{subfigure}[t]{.15\columnwidth}
	\centering
	\includegraphics[width = 1\columnwidth]{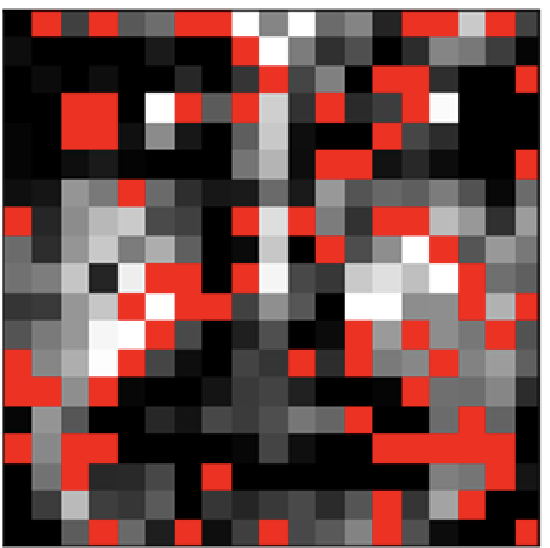}
	\subcaption{}
\end{subfigure}
\hfill
\begin{subfigure}[t]{.15\columnwidth}
	\centering
	\includegraphics[width = 1\columnwidth]{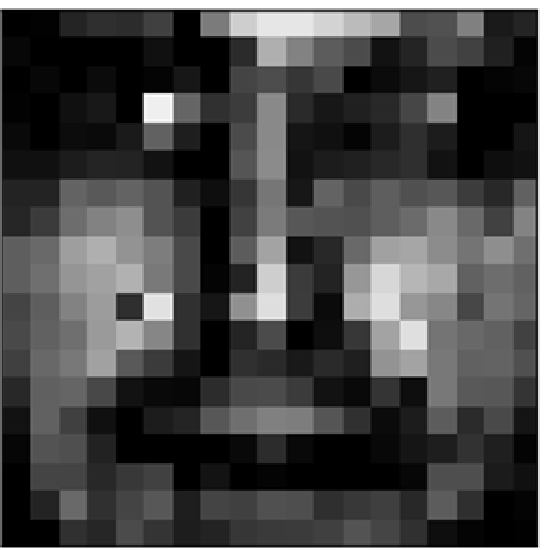}
	\subcaption{}
\end{subfigure}
\hfill
\begin{subfigure}[t]{.15\columnwidth}
	\centering
	\includegraphics[width = 1\columnwidth]{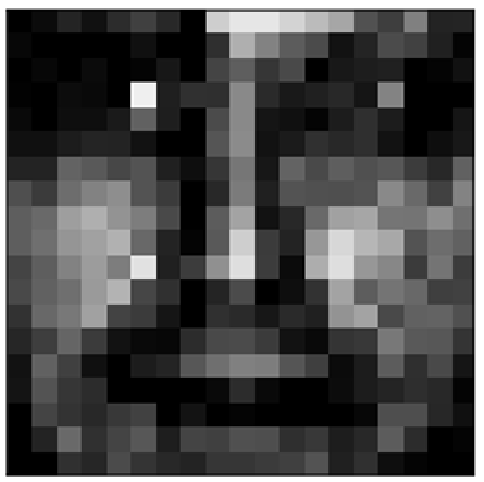}
	\subcaption{}
\end{subfigure}
	
\caption{(a) Original image $\bV$; (b) Masked training image (missing pixels in red) $\bV$; (c)~Adversary's added-on masked image $\bR^*$; (d)~Adversarially trained masked image $\bV + \bR^*$; (e)~Restored image using AT-NMF; (f)~Restored image using NMF when $\alpha = 0.2$ and  $\lambda = 2$. }
\label{fig:restored}
\end{figure}

\begin{figure}[hbt!]
\centering
\begin{tabular}{ccc}
	\includegraphics[width = .29\columnwidth]{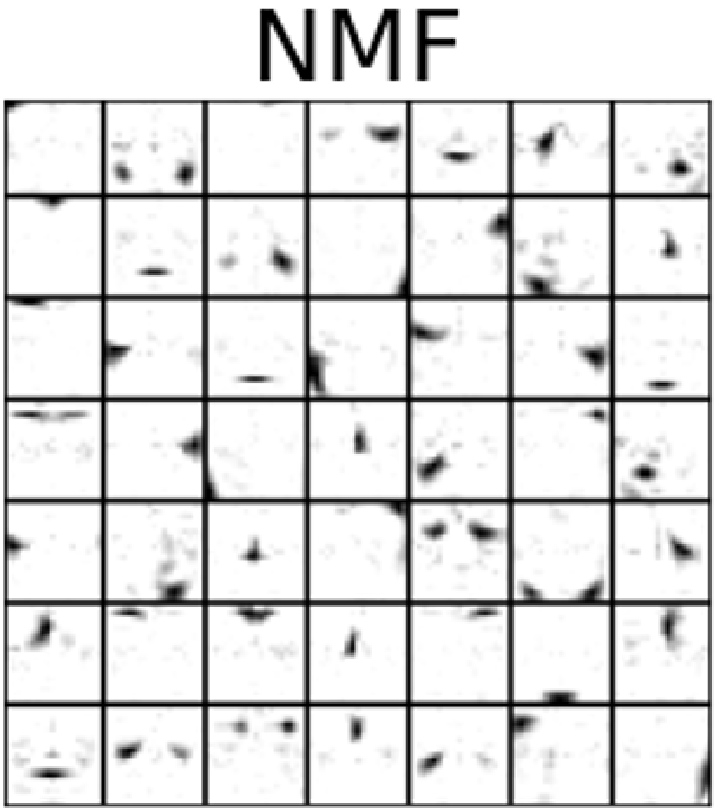} & \includegraphics[width = .29\columnwidth]{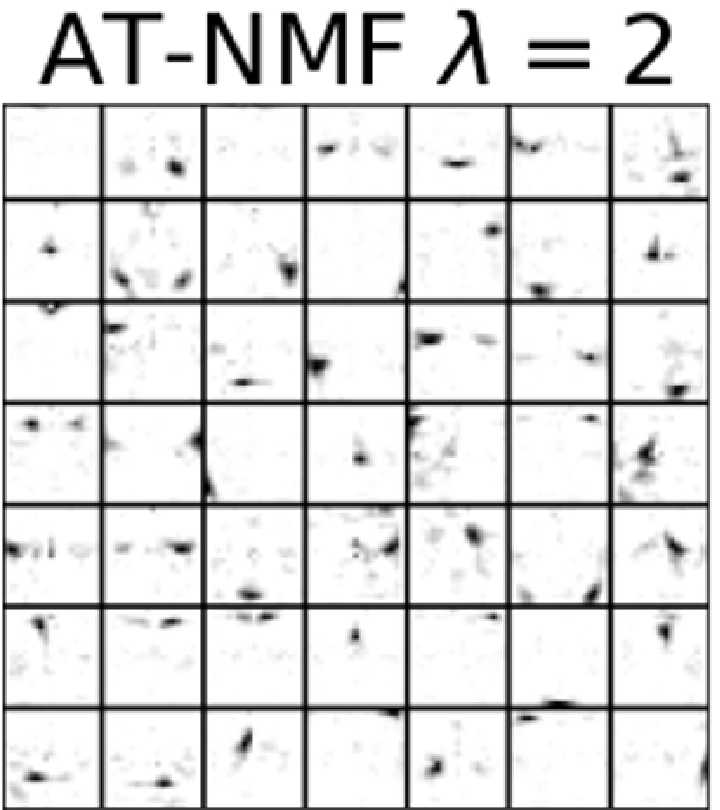} & \includegraphics[width = .29\columnwidth]{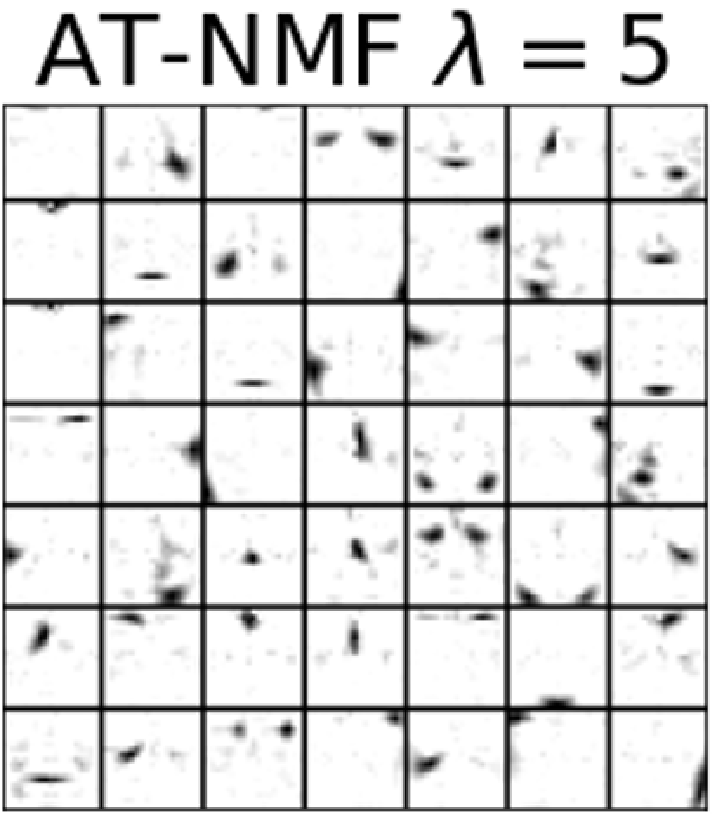}  
\end{tabular}
\caption{Parts learnt on CBCL dataset when $\alpha = 0.1$}
\label{fig:CBCL Parts}
\end{figure}

\subsection{Simulations on the CBCL Dataset} \label{sec:cbcl}
We report experiments on the CBCL dataset~\cite{LS99}. This dataset consists of $N=2429$ frontal-view facial images, each having $F=19 \times 19 = 361$ pixels. Thus, $\bV \in \bbR_+^{361 \times 2429}$. We normalized each image using the procedure outlined in~\cite{LS99}.  Table~\ref{tab:CBCL} displays the results. We observe that AT-NMF with $\lambda=2$  outperforms all competing algorithms.  The top right plot in Fig.~\ref{fig:rmse_lambda} also indicates that the performance of AT-NMF converges to that of NMF as  $\lambda\to\infty$.

Fig.~\ref{fig:CBCL Loss} shows the evolution of the objective function of standard NMF and AT-NMF  with $\lambda = 2$ and $\lambda = 5$ for the {\em outer} iterations of Algorithm~\ref{alg:atnmf}. For NMF, as expected, the objective function values converge.  For AT-NMF, we observe that for small $\lambda$, the iterates  exhibit long-term oscillatory behavior. This is because the MM property that guarantees that the loss is monotonically non-increasing is not present in AT-NMF where we maximize over $\bR$ and minimize over ($\bW$, $\bH$). We leave the convergence  of AT-NMF to future work. 
Notice that despite reaching a higher objective value than NMF, the prediction performance is better. This is due to the built-in generalization ability of adversarial training in AT-NMF. 


Fig.~\ref{fig:restored} shows an example of a masked training image and the corresponding image restored  using AT-NMF. In AT-NMF, adding $\bR^*$ on  $\bV$ intensifies the differences between the pixels, which results in a ``training'' image $\bV + \bR^*$ (subfigure~(d)) whose features (eyes, nose, lower cheeks) become more distinctive. 
We observe that the restored image by AT-NMF (subfigure~(e)) is similar to the original image (subfigure~(a)). 



Fig.~\ref{fig:CBCL Parts} shows the parts learned. These are the columns of the   basis matrix $\bW$.  We observe that AT-NMF retains the ability to extract parts-based representations. Specifically, the parts learned by AT-NMF with $\lambda = 2$ contain   more noise in each basis image. This is natural due to the effect of adding~$\bR^*$.

\vspace{-.05in}

\subsection{Simulations on   Hyperspectral Datasets}

\begin{table}[t!]
	\caption{RMSE of the Moffet Dataset}
	\label{tab:moffet}
	\centering
	\begin{adjustbox}{width=0.5\textwidth}
		\begin{tabular}{ |c|c|c|c|c| } 
			\hline
			$\alpha$ & \textbf{NMF} & \textbf{ANMF} & \textbf{AT-NMF} ($2$)  & \textbf{AT-NMF} ($5$)  \\
			\hline
			$0.3$ & $0.056 \pm 0.000$ & $0.050 \pm 0.000$ & $\mathbf{0.041} \pm \mathbf{0.001}$ & $0.049 \pm 0.000$  \\ 
			\hline
			$0.4$ & $0.072 \pm 0.000$ & $0.067 \pm 0.000$ & $\mathbf{0.052} \pm \mathbf{0.001}$ & $0.064 \pm 0.000$  \\ 
			\hline
			$0.5$ & $0.088 \pm 0.000$ & $0.084 \pm 0.000$ & $\mathbf{0.064} \pm \mathbf{0.000}$ & $0.079 \pm 0.000$  \\ 
			\hline
			$0.6$ & $0.103 \pm 0.000$ & $0.100 \pm 0.000$ & $\mathbf{0.079} \pm \mathbf{0.000}$ & $0.095 \pm 0.000$  \\ 
			\hline
			$0.7$ & $0.120 \pm 0.000$ & $0.117 \pm 0.000$ & $\mathbf{0.097} \pm \mathbf{0.000}$ & $0.113 \pm 0.000$  \\ 
			\hline
			$0.8$ & $0.136 \pm 0.000$ & $0.134 \pm 0.000$ & $\mathbf{0.117} \pm \mathbf{0.000}$ & $0.131 \pm 0.000$  \\ 
			\hline
			$0.9$ & $0.151 \pm 0.000$ & $0.150 \pm 0.000$ & $\mathbf{0.140} \pm \mathbf{0.000}$ & $0.148 \pm 0.000$  \\ 
			\hline
		\end{tabular}
	\end{adjustbox}
\end{table}

\begin{table}[t!]
	\caption{RMSE of the Madonna dataset}
	\label{tab:Madonna}
	\centering
	\begin{adjustbox}{width=0.5\textwidth}
		\begin{tabular}{ |c|c|c|c|c| } 
			\hline
			$\alpha$ & \textbf{NMF} & \textbf{ANMF} & \textbf{AT-NMF} ($2$)  & \textbf{AT-NMF} ($5$) \\
			\hline
			$0.3$ & $0.120 \pm 0.000$ & $0.106 \pm 0.000$ & $\mathbf{0.103} \pm \mathbf{0.006}$ & $0.109 \pm 0.000$  \\ 
			\hline
			$0.4$ & $0.147 \pm 0.000$ & $0.139 \pm 0.000$ & $\mathbf{0.117} \pm \mathbf{0.004}$ & $0.134 \pm 0.000$  \\ 
			\hline
			$0.5$ & $0.175 \pm 0.000$ & $0.170 \pm 0.000$ & $\mathbf{0.131} \pm \mathbf{0.006}$ & $0.161 \pm 0.000$  \\ 
			\hline
			$0.6$ & $0.204 \pm 0.000$ & $0.201 \pm 0.000$ & $\mathbf{0.158} \pm \mathbf{0.003}$ & $0.190 \pm 0.000$  \\ 
			\hline
			$0.7$ & $0.232 \pm 0.000$ & $0.229 \pm 0.000$ & $\mathbf{0.185} \pm \mathbf{0.005}$ & $0.219 \pm 0.000$  \\ 
			\hline
			$0.8$ & $0.261 \pm 0.000$ & $0.260 \pm 0.000$ & $\mathbf{0.227} \pm \mathbf{0.002}$ & $0.251 \pm 0.000$  \\ 
			\hline
			$0.9$ & $0.290 \pm 0.000$ & $0.289 \pm 0.000$ & $\mathbf{0.271} \pm \mathbf{0.000}$ & $0.284 \pm 0.000$  \\ 
			\hline
		\end{tabular}
	\end{adjustbox}
\end{table}

Finally, we describe experiments on two hyperspectral  datasets, namely the Moffet~\cite{Moffet} and Madonna~\cite{Madonna}  datasets, which are normalized in the same way as in~\cite{Fevotte_2015}. This leads to $F=165$ and $F=160$ spectral bands in the Moffet and Madonna datasets respectively. The size of the images (number of pixels) is $50\times 50$ so $N=2500$. We set $K = 5$. 

Tables~\ref{tab:moffet} and~\ref{tab:Madonna} display the results for these two datasets.  We observe that AT-NMF with $\lambda = 2$ outperform all other  algorithms for all $\alpha$.  The bottom plots in Fig.~\ref{fig:rmse_lambda} also indicate that as $\lambda \to \infty$, AT-NMF reverts to standard NMF.


\section{Conclusion}
We considered a formulation of AT-NMF in which an adversary adds a matrix $\bR$ of bounded norm to the data matrix $\bV$. We designed an efficient algorithm  to solve the min-max problem. This algorithm ``shapes'' $\bR$ in a way such that the Frobenius norm of $\bV+\bR$ and the approximating matrix $\bW\bH$ is maximized. We show, through extensive numerical experiments, that AT-NMF outperforms standard NMF and   ANMF~\cite{pmlr-v119-luo20c} in terms of prediction. Designing algorithms for other divergence measures such as $\beta$-divergences~\cite{fevotte2009nonnegative}, different bounded sets~$\calR$ and online implementations of  NMF~\cite{ZhaoTan17} would be interesting.  Quantifying the generalization ability is also an avenue for future  theoretical research.


\newpage

\bibliographystyle{unsrt}
\bibliography{biblio}


\end{document}

%% file: preamble.tex
\catcode`~=11 \def\UrlSpecials{\do\~{\kern -.15em\lower .7ex\hbox{~}\kern .04em}} \catcode`~=13 

\allowdisplaybreaks[1]

\newcommand{\calR}{\mathcal{R}}

\newcommand{\bA}{\mathbf{A}}

\newcommand{\bh}{\mathbf{h}}
\newcommand{\bH}{\mathbf{H}}

\newcommand{\bj}{\mathbf{j}}

\newcommand{\bR}{\mathbf{R}}

\newcommand{\bU}{\mathbf{U}}

\newcommand{\bV}{\mathbf{V}}

\newcommand{\bW}{\mathbf{W}}

\newcommand{\bZ}{\mathbf{Z}}

\newcommand{\rmF}{\mathrm{F}}

\newcommand{\bbR}{\mathbb{R}}

\DeclareMathAlphabet{\mathbsf}{OT1}{cmss}{bx}{n}
\DeclareMathAlphabet{\mathssf}{OT1}{cmss}{m}{sl}

\DeclareSymbolFont{bsfletters}{OT1}{cmss}{bx}{n}  
\DeclareSymbolFont{ssfletters}{OT1}{cmss}{m}{n}
\DeclareMathSymbol{\bsfGamma}{0}{bsfletters}{'000}
\DeclareMathSymbol{\ssfGamma}{0}{ssfletters}{'000}
\DeclareMathSymbol{\bsfDelta}{0}{bsfletters}{'001}
\DeclareMathSymbol{\ssfDelta}{0}{ssfletters}{'001}
\DeclareMathSymbol{\bsfTheta}{0}{bsfletters}{'002}
\DeclareMathSymbol{\ssfTheta}{0}{ssfletters}{'002}
\DeclareMathSymbol{\bsfLambda}{0}{bsfletters}{'003}
\DeclareMathSymbol{\ssfLambda}{0}{ssfletters}{'003}
\DeclareMathSymbol{\bsfXi}{0}{bsfletters}{'004}
\DeclareMathSymbol{\ssfXi}{0}{ssfletters}{'004}
\DeclareMathSymbol{\bsfPi}{0}{bsfletters}{'005}
\DeclareMathSymbol{\ssfPi}{0}{ssfletters}{'005}
\DeclareMathSymbol{\bsfSigma}{0}{bsfletters}{'006}
\DeclareMathSymbol{\ssfSigma}{0}{ssfletters}{'006}
\DeclareMathSymbol{\bsfUpsilon}{0}{bsfletters}{'007}
\DeclareMathSymbol{\ssfUpsilon}{0}{ssfletters}{'007}
\DeclareMathSymbol{\bsfPhi}{0}{bsfletters}{'010}
\DeclareMathSymbol{\ssfPhi}{0}{ssfletters}{'010}
\DeclareMathSymbol{\bsfPsi}{0}{bsfletters}{'011}
\DeclareMathSymbol{\ssfPsi}{0}{ssfletters}{'011}
\DeclareMathSymbol{\bsfOmega}{0}{bsfletters}{'012}
\DeclareMathSymbol{\ssfOmega}{0}{ssfletters}{'012}

\newcommand{\tilh}{\tilde{h}}

\newcommand{\tilbh}{\tilde{\bh}}

\newcommand{\hatv}{\hat{v}}

\newcommand{\hatbV}{\hat{\bV}}

\newcommand{\tilbW}{\tilde{\bW}}

\newcommand{\bzero}{\mathbf{0}}

\newtheorem{definition}{Definition}

\newcommand{\qednew}{\nobreak \ifvmode \relax \else
      \ifdim\lastskip<1.5em \hskip-\lastskip
      \hskip1.5em plus0em minus0.5em \fi \nobreak
      \vrule height0.75em width0.5em depth0.25em\fi}